\pdfoutput=1

\documentclass[11pt]{article}

\usepackage{acl}

\usepackage{times}
\usepackage{latexsym}
\usepackage{amsmath}

\usepackage{graphicx}
\usepackage{booktabs}
\usepackage{graphicx}
\usepackage{multirow}
\usepackage{array}
\usepackage{arydshln} 
\usepackage{float}

\usepackage[medium,compact]{titlesec}

\usepackage[T1]{fontenc}

\usepackage[utf8]{inputenc}

\usepackage{microtype}

\title{Syntax Controlled Knowledge Graph-to-Text Generation with Order and Semantic Consistency}

\author{Jin Liu$^1$, Chongfeng Fan$^1$, Fengyu Zhou$^1$, Huijuan Xu$^2$ \\
       $^1$ Shandong University \\
       $^2$ Pennsylvania State University\\
        \{\texttt{202120638,202114753,zhoufengyu}\}
        \texttt{@mail.sdu.edu.cn,} \\
        \texttt{hkx5063@psu.edu}
        }
        
\begin{document}
	
	\maketitle
	
	\begin{abstract}
		The knowledge graph (KG) stores a large amount of structural knowledge, while it is not easy for direct human understanding. Knowledge graph-to-text (KG-to-text) generation aims to generate easy-to-understand sentences from the KG, and at the same time, maintains semantic consistency between generated sentences and the KG. Existing KG-to-text generation methods phrase this task as a sequence-to-sequence generation task with linearized KG as input and consider the consistency issue of the generated texts and KG through a simple selection between decoded sentence word and KG node word at each time step. However, the linearized KG order is commonly obtained through a heuristic search without data-driven optimization. In this paper, we optimize the knowledge description order prediction under the order supervision extracted from the caption and further enhance the consistency of the generated sentences and KG through syntactic and semantic regularization. We incorporate the Part-of-Speech (POS) syntactic tags to constrain the positions to copy words from the KG and employ a semantic context scoring function to evaluate the semantic fitness for each word in its local context when decoding each word in the generated sentence. Extensive experiments are conducted on two datasets, WebNLG and DART, and achieve state-of-the-art performances. Our code is now public available\footnote{\url{https://github.com/LemonQC/KG2Text}}.
	\end{abstract}
	\section{Introduction}
\begin{figure}
		\centering
		\includegraphics[scale=0.45]{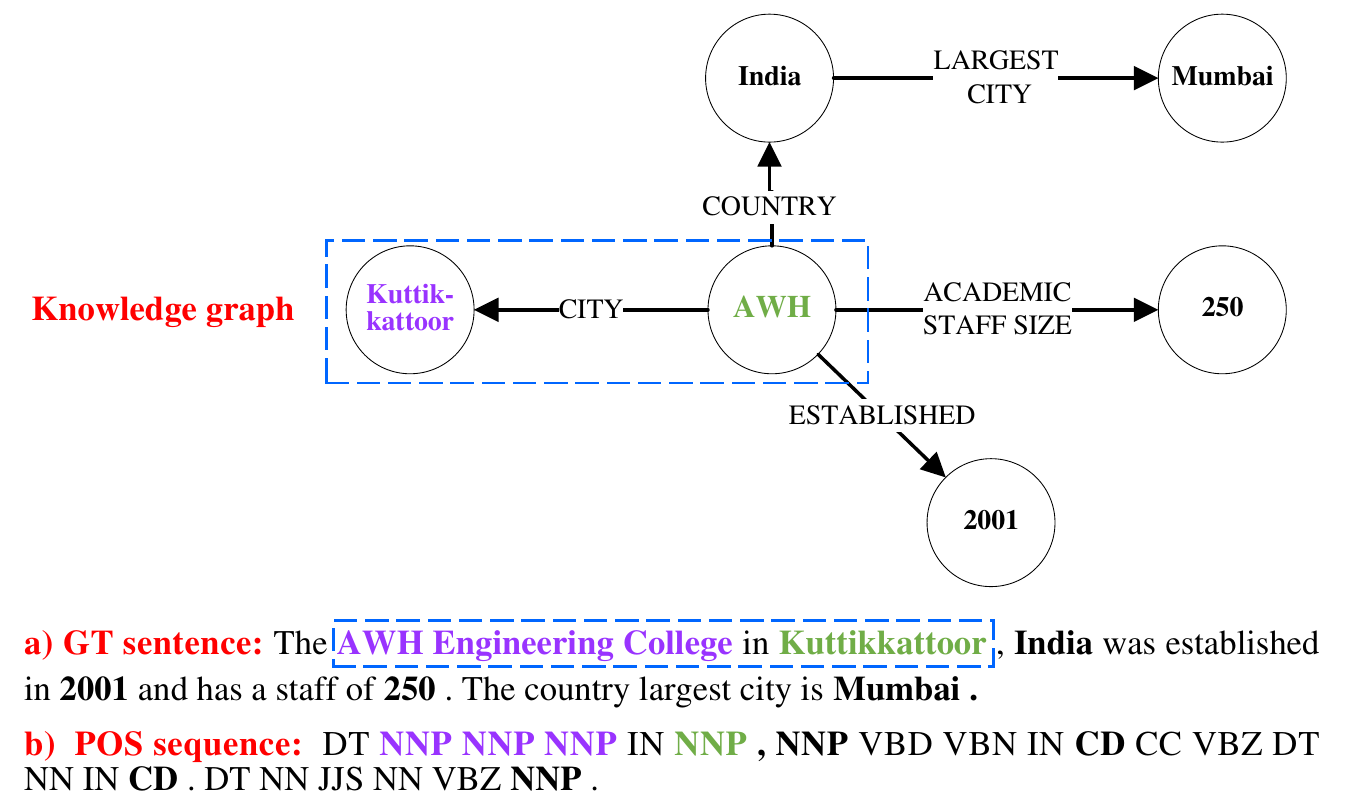}
		\caption{One example KG from DART dataset~\cite{dartdataset}. AWH is short for AWH Engineering College. Text a) is the ground-truth sentence, text b) is the POS sequence obtained by NLTK, and most words copied from the KG are nouns. The linearized order of KG should correlate with the information sequence expressed in the corresponding sentence (words indicated in bold in GT sentence), and each decoded word should fit in its local semantic context (e.g. the blue dashed lines).
		}
		\label{fig:figure01}
	\end{figure}

	Knowledge graphs (KGs) record the common sense knowledge in a structural way and have many potential applications, e.g., question answering~\cite{paper0001}, recommendation system~\cite{paper0002} and storytelling~\cite{paper0003}. 
	One typical KG is shown in Figure~\ref{fig:figure01}, with circle nodes indicating entities, and the directional edges connecting the head node to the tail node and representing the relation among connected entities. This structural representation in KG is easy for information storage while not convenient for human understanding. 
	In this paper, we focus on the task of KG-to-text generation, which aims to describe an input KG with fluent language sentences in an easy-to-understand way. Compared to the traditional text generation task, KG-to-text generation poses the extra challenge of maintaining the word authenticity in the generated sentence given the input KG.
With the word authenticity and sentence fluentness in mind, existing KG-to-text generation methods~\cite{plms,graphtrans,webnlg} phrase this task as a sequence-to-sequence generation task, where the KG is linearized as input sequence and decoded into sentences, and the word from the KG are selected to be inserted into the decoded sentence with predicted confidence at each time step. However, when the KG is linearized into a sequence, simple heuristic search-based algorithms are usually utilized, e.g., breadth-first search (BFS) or using other pre-defined rules for sorting~\cite{fewshot,plms}, without considering the word sequence information in the ground-truth sentences. The order inference of the KG is not tightly correlated to the word sequence information in the ground-truth sentences and is generated in a disjoint prestage. The decoded sentence is further conditioned on the linearized KG order which might incur cascaded errors~\cite{paper0006}.
To tackle this problem, we extract the order information from the ground-truth sentence and use this order information to directly supervise the KG order prediction with graph structural local context. In this case, our order prediction component for KG encodes the sentence sequence prior information and will benefit the sentence generation in the follow-up stage.

Also, most existing methods~\cite{fewshot,graphtrans} maintain the word authenticity by maximizing the copy probability of the tokens from the KG, while ignoring the syntactic correctness and semantic relevance. A simple observation from the example in Figure~\ref{fig:figure01}, you can find that the POS tags of most words copied from the KG are nouns. Motivated by this observation, we introduce a POS generator to guide the sentence generation process by applying the POS information as additional supervision at each time step and limit the position scope of the word selection from KG. Moreover, to further enhance semantic relevance of generated sentence, we consider the structural information and local semantic information of the KG by designing a semantic context scoring function with sliding windows of different sizes, and combine the semantic context score into the word selection process at each time step of the sentence generation.

In summary, we propose a Syntax controlled KG-to-text generation model with Order and Semantic Consistency, called S-OSC. 
The main contributions are summarized as follows:
	\begin{itemize}
		\item We propose a learning-based sorting network to obtain the optimal KG description order with graph structural context for more fluent caption generation.
		\item We enhance the authenticity of generated sentences to the KG through syntactic and semantic regularization. POS tag information is incorporated into the sentence modeling and helps to determine the word selection from KG, together with one additional semantic context scoring function.
		\item Extensive results on two benchmark datasets indicate that our proposed S-OSC model outperforms previous models and achieves new state-of-the-art performance.
	\end{itemize}
	
\section{Related Work}
	\subsection{KG-To-Text Generation}
	KG-to-text generation task has been a hot research topic since the first dataset WebNLG was proposed~\cite{webnlg}. Recent works for solving this task have two main categories. One category is using graph neural networks~\cite{paper0009,globallocal,fewshot,cheng2020ent} or graph transformers~\cite{graphtrans} to directly capture the graph structure information and decode into sentences. E.g., the recent work~\cite{globallocal} forms four different encoder architectures for combining local and global node contexts. ENT-DESC~\cite{cheng2020ent} introduces multi-graph structure to better aggregate knowledge information. The other category is first linearizing the KG~\cite{plms,paper0013,webnlg,hoyle2020promoting} and then formulating a sequence-to-sequence generation task with lineared KG nodes as input to generate sentences. E.g., Distiawan et al.~\cite{paper0008} utilize a fixed tree traversal order to directly flatten the KG into a linearized representation. The structural information of the KG is not preserved when generating the KG order. In this paper, we follow the second general pipeline with local structure encoded in the order generation process.
	\subsection{Knowledge Graph Order Generation}
	A line of KG-to-text models~\cite{plms,paper0011} tries to generate sentences conditioned on the KG order, where the order generation is especially important as different KG description orders may result in various generated texts. Most previous works focus on graph traversal-based approaches~\cite{paper0012,webnlg,jointgt,fewshot} for KG order generation.~\cite{fewshot} proposes to use a relation-biased breadth first search (RBFS) strategy to linearize the KG. These previous graph traversal-based approaches are heuristic without considering the word sequence information in the ground-truth sentences. Inspired by previous work~\cite{paper0006} which generates image caption with optimal object description order, we extract the sequence information from the ground-truth sentences as supervision and train one order prediction module to generate optimal order. Moreover, our order prediction considers the local graph structure in triplet.
	\subsection{Captioning with POS Tags}
	POS tags have been used in various text generation (e.g. image captioning and video captioning) to impose the syntactic constraint. In the neural text generation work~\cite{pos0003}, the authors propose to use POS guided softmax function as the linguistic prior information for modeling the posterior probabilities of next-POS and next-token, in order to increase text generation diversity. In the image caption, Bugliarello et al.~\cite{pos0002} claim that incorporating POS tag information in the sentence generation process consistently improves the quality of the generated text. In video captioning, Hou et al.~\cite{pos0001} propose to define the templates of POS tag sequences to represent the syntactic structure of the generated text. In KG-to-text generation task, we not only use the POS tags to ensure the syntactic correctness of the generated text, but also use the POS tags to constrain the positions to copy words from KGs.
	\subsection{Pre-Trained Language Models}
    Pre-trained language models (PLMs) on massive corpora, such as BERT~\cite{bert}, BART~\cite{bart} and T5~\cite{t5}, have achieved superior performance in various natural language generation tasks, including KG-to-text generation task~\cite{plms,paper0010,jointgt}. Ribeiro et al.~\cite{plms} leverage the generation ability of PLMs and use the linearized KG as input to generate texts. We also use the PLM in our method to guarantee the model generalization ability, and at the same time, design extra order prediction and context scoring components to maintain the semantic consistency.
\section{Approaches}
	In this section, we first formulate the KG-to-text generation problem setting and then elaborate the proposed S-OSC in detail.
	\subsection{Problem Formulation}
	Given the input KG $\mathcal{G}$, which is composed of $\{(h_1, r_1, t_1), \cdots,  (h_n, r_n, t_n)|h_*, t_*\in \mathcal{E}, r_*\in \mathcal{R}$\}, where $\mathcal{E}$ denotes the entity set and $\mathcal{R}$ represents the relation set, the KG-to-text generation task aims to generate a fluent and reasonable text sequence $\mathcal{T}=<t_1,t_2,\cdots,t_k>(t_k\in \mathcal{V})$, where $\mathcal{V}$ denotes the vocabulary. In this paper, we follow the general pipeline~\cite{plms,jointgt,paper0005} of linearizing the input KG into sequence $\mathcal{G}_{linear}=<g_1,g_2,\cdots,g_m>$ consisting of $m$ tokens, and then decoding the KG token sequence into sentences.
	\subsection{Our Proposed S-OSC Model}
\begin{figure*}
	    \centering
		\includegraphics[scale=0.65]{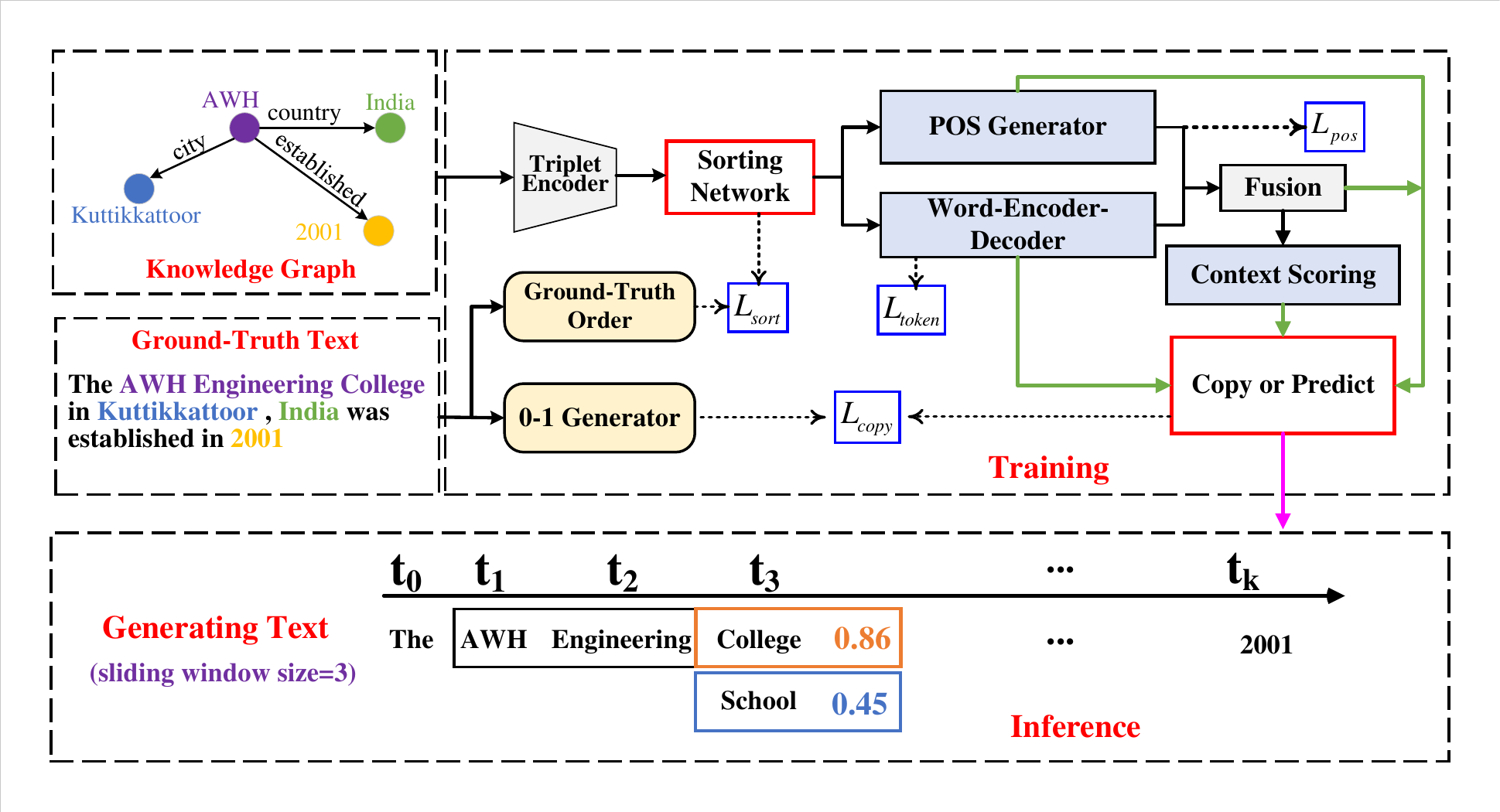}
		\caption{S-OSC model architecture.
		The pre-trained language model takes the KG as input and extracts head-relation-tail triplet structure features through Triplet Encoder. 
		Then, these triplet features are fed into the Sorting Network to generate one optimal description order under the sequence supervision extracted from ground-truth sentence.
		Conditioned on the KG order, we further
		decode the sequence into sentences through Word Decoder and apply additional syntactic supervision through POS Generator.
		To maintain the word authenticity in the generated sentences, we further design one Copy or Prediction component incorporating the POS syntactic information and semantic context scoring with sliding windows to help determine when to copy tokens from KG, in addition to the word decoding probability from the Word-Encoder-Decoder and the binary classification probability for copy or prediction at each time step. 
		}
		\label{fig:arch}
	\end{figure*}

We propose S-OSC, illustrated in Figure~\ref{fig:arch}, which consists of two main components: one learning-based sorting network for KG, and one copy or prediction selection module for decoding each word in the sentences. The sorting network generates the optimal description sequence for the input KG. Based on the KG order sequence, the sentence decoder generates each word with a certain probability predicted by the copy or prediction selection module to replace the decoded word with the word in the KG. Thus, the model can maintain the word authenticity in the generated sentence compared to the KG. A key innovation in our learning-based sorting network is to utilize the sequence information extracted from the ground-truth sentence to directly supervise the optimal sequence prediction instead of heuristic search in the KG without considering the description sequence prior. Our copy or prediction selection module for decoding each word in the sentences incorporates additional POS syntactic constraint and semantic context consistency scoring function evaluating the semantic fitness of each word in its sliding windows with various sizes. The details of each module in our S-OSC model are illustrated in the following sections.
\subsubsection{Sorting Network}
	The description order of the KG will affect the content of the generated sentence. In a worse scenario, poor order may result in the loss of important information (see the example in Figure~\ref{fig:quality}). To overcome the drawback of disjoint learning for KG order generation and sentence generation in the previous heuristic-based methods~\cite{fewshot,plms,paper0013}, we propose a learning-based sorting network with the order supervision extracted from the ground-truth sentence. Notably, our sorting network is based on the features from the structural Triplet Encoder, where the head-relation-tail triplet structure features are extracted through pre-trained KG embedding method TransH~\cite{transH} and pre-trained language model BART~\cite{bart}. Due to the variable length of the KG, we introduce a placeholder to pad it into fixed length $N$, which also denotes the number of possible position classes. The head-relation-tail triplet structure features $F_{stru}$ are concatenated with the padding $F_{pad}$ and fed through the Fully Connected (FC) layers with softmax classifier $FC_{s}$ to obtain $S_{matrix}$ and predict the sorting order $S_{order}$.
\begin{equation}
	\begin{split}
	   S_{matrix}=\text{FC}_{s}([F_{stru};F_{pad}])\\
	  S_{order}=\text{argmax}_{row}(S_{matrix})   
	\end{split}
	\end{equation}
In this paper, we treat the order prediction task as a classification problem, where N denotes the number of classes (the maximum of the triplets in KG). Thus, we measure the cross-entropy loss between the ground-truth order $G_{order}$ and the sorting order $S_{order}$:
\begin{equation}\label{cross}
		L_{sort}=-\sum_{n=0}^{N} \text{log}(S_{matrix}^n)\cdot G_{order}^n
\end{equation}
\begin{figure}[H]
	\centering
	\includegraphics[scale=0.8]{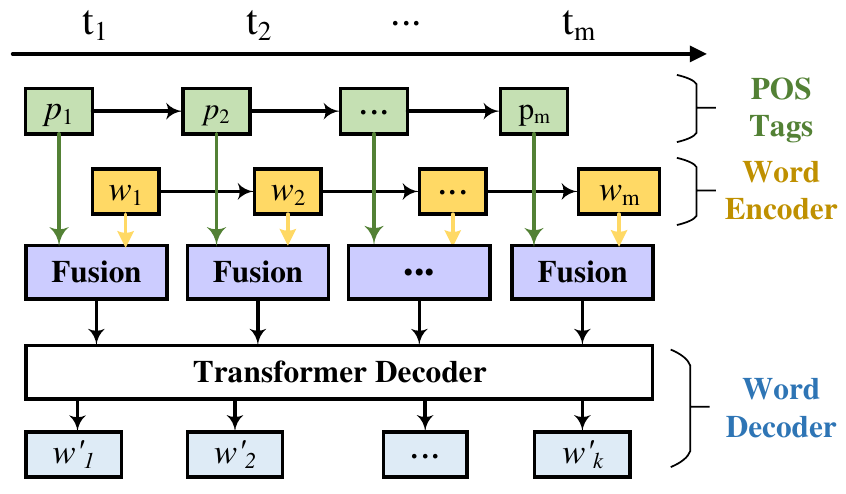}
	\caption{The architecture of the Word-Encoder-Decoder with a POS Generator.}
	\label{fig:posgen}
\end{figure} 
\subsubsection{Copy or Prediction Network}
\begin{figure}
		\centering
		\includegraphics[scale=0.8]{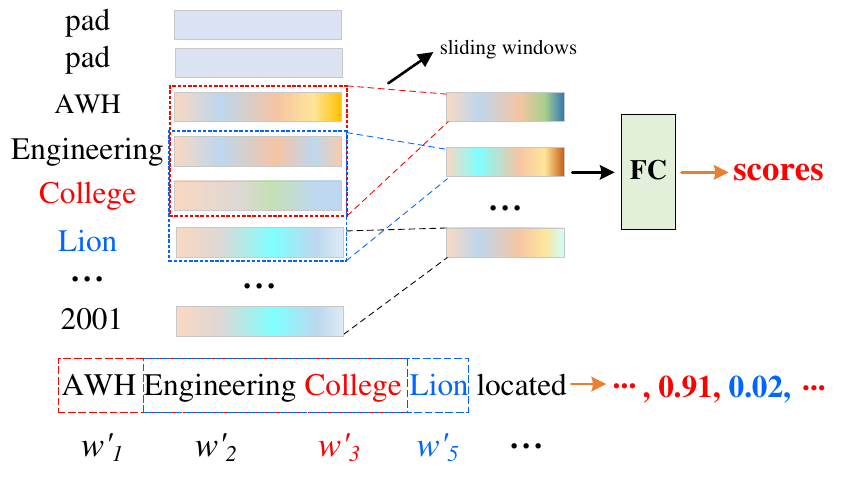}
		\caption{The architecture of semantic context scoring module with sliding windows.}
		\label{fig:contextsemantic2}
\end{figure}

To maintain the authenticity of KG words in generated sentences, the model needs to selectively copy words from KG words instead of using the predicted words from Word Decoder. Besides directly predicting the copy probability from the hidden state~\cite{graphtrans,fewshot,copypaste}, the Copy or Prediction Network in our S-OSC model further enhances the syntactic and semantic consistency of selected words in the decoded sentence through the incorporation of POS generator (shown in Figure~\ref{fig:posgen}) and semantic context scoring (shown in Figure~\ref{fig:contextsemantic2}). Next, we will introduce each module in detail. 
\paragraph{POS Syntactic Constraint}\label{posgenerator}
Conditioned on the KG order $G_{order}$, we first linearize the KG by adding the tokens $<Head>, <Relation>, <Tail>$ to the corresponding position for each triplet and obtain the $\mathcal{G}_{linear}$. Then, the Word Encoder and the POS Generator take $G_{linear}$ as their inputs and output the word encoding WI=$\{w_{i}, i\in 1\cdots m\}$ and POS tag encoding PI=$\{p_{i}, i\in 1\cdots m\}$ , respectively. Then, the token encoding $w_i$ and POS tag encoding $p_i$ are combined in the fusion module to get the updated token encoding $w_i$.
\begin{equation}
    w_i =\text{LN}(FC([w_i;p_i])+w_i),
\end{equation}
where LN denotes the layer normalization. The updated token encoding $w_i$ after fusing is decoded into sentence $\text{WI}^{'}=\{w_{i}^{'}, i\in 1\cdots k\}$ in Word Decoder.

POS generator is supervised through POS tags pre-extracted from the sentence. The loss function is formulated as:
\begin{equation}\label{pos}
	L_{pos}=-\sum_{l=1}^{M}\text{log}(P_{gen}(p_l|p_1,\cdots,p_{l-1};G_{order})),
\end{equation}
where $P_{gen}$ denotes the predicted probability from POS generator. Similarly, the objective of Word-Encoder-Decoder is as follows:
\begin{equation}\label{word}
	L_{token}=-\sum_{j=1}^{K}\text{log}(W_{gen}(w_j^{'}|w_1^{'},\cdots,w_{j-1}^{'};\text{WI}^{'})),
\end{equation} 
where $W_{gen}$ denotes the predicted probability of each word token. 
\paragraph{Semantic Context Scoring}
Besides the syntactic constraint for copied words, we also design one semantic context scoring component, illustrated in Figure~\ref{fig:contextsemantic2}, to evaluate the semantic consistency of copied or predicted words in the sliding windows. 
Sliding windows are generated for each word to provide the local context, e.g., the sliding window size is set to 3 in Figure~\ref{fig:contextsemantic2}. Besides, padding is needed for the first several words when forming the sliding windows.
Word features in the sliding window are contacted to get the context information $F_{context}$, and are fed into the FC layers to obtain the semantic score $X_{semantic}$.
	\begin{equation}
		X_{semantic}=\sigma (\text{FC}(F_{context})),
	\end{equation}
where $\sigma$ denotes the sigmoid function.
\paragraph{Word Copy Probability Prediction}
With our newly introduced POS token embeddings $v_{p_k}$ and the semantic context score X$_{semantic}$, the probability $p_{copy}^k$ for copying words from KG is computed in Eq.~\ref{followeq}, and is used in testing time for final selection between predicted words from Word Decoder and words in KG at each time step when generating sentences.
\begin{equation}
    \begin{split}
        t_{copy}^k=\sigma (W_{1}v_{w_k}+W_{2}v_{p_k}+W_{3}s_k+b_{copy}),\\
        p_{copy}^k=\lambda \cdot X_{semantic} + (1-\lambda) \cdot t_{copy}^{k},
    \end{split}
    \label{followeq}
\end{equation}
where $W_1$, $W_2$, $W_3$ and $b_{copy}$ are learnable parameters. $v_{w_k}$ represents token embedding and $s_k$ represents the last hidden state of Word-Decoder at each time step. $\lambda$ is a trade-off coefficient and is set as 0.3.

The semantic context scoring module is jointly optimized with copy probability prediction and benefits the copy probability prediction. The copy or prediction loss function is defined as:
\begin{equation}\label{copy}
		\begin{aligned}
			L_{copy}=-\sum_{k=0}^{K}(y^k\cdot \text{log}(p_{copy}^k)+(1-y^k)\\
			\cdot \text{log}(1-p_{copy}^k)),
		\end{aligned}
\end{equation}

where $y^k$ is the ground-truth 0-1 label indicating copying or predicting word at $k-th$ time step, which is generated from KG and the ground-truth sentence (see more details in supplementary material). 

Finally, the total training loss $L_{total}$ in our S-OSC model is composed of four components: sorting loss $L_{sort}$(Eq.~\ref{cross}), POS generation loss $L_{pos}$(Eq.~\ref{pos}), word generation loss $L_{token}$(Eq.~\ref{word}) and copy or prediction loss $L_{copy}$(Eq.~\ref{copy}).
\begin{equation}\label{totalloss}
	L_{total}=L_{token}+\lambda_1 L_{pos}+\lambda_2 L_{sort} +\lambda_3 L_{copy},
\end{equation}
where $\lambda_1$, $\lambda_2$ and $\lambda_3$ are the trade-off coefficients.

\section{Experiments}
In this section, we report the comparison results with state-of-the-art methods and further analyze the performance of each component in our S-OSC model through ablation studies. We also evaluate our model performance through human evaluation and qualitative analysis.
\subsection{Datasets}
In this paper, two benchmarks: WebNLG~\cite{fewshot} and DART~\cite{dartdataset} are utilized to evaluate the performance of our S-OSC model.
\paragraph{WebNLG} 
WebNLG~\cite{fewshot} is a most widely used dataset in KG-to-text generation task. Each graph is extracted from DBPedia and consists of two to seven triplets. The train/val/test splits are 7362/1389/5427.
\paragraph{DART}
Compared to WebNLG, DART~\cite{dartdataset} is a larger open-domain dataset, where triples are composed of tree-structured ontology. The train/val/test splits are 30348/2759/5097.
\subsection{Evaluation Metrics}
Following the previous works~\cite{plms,globallocal,fewshot} on WebNLG dataset, we adopt four automatic language evaluation metrics for WebNLG dataset, i.e., BLEU-4~\cite{bleu}, CIDEr~\cite{cider}, Chrf++~\cite{chrf} and ROUGE-L~\cite{rouge}. Following previous works~\cite{dartdataset} on DART dataset, in addition to BLEU-4~\cite{bleu}, we use four additional automatic evaluation metrics, i.e., METEOR~\cite{meteor}, MoverScore~\cite{moverscore}, BERTScore~\cite{bertscore} and BLEURT~\cite{bleurt}.
\subsection{Implementation Details}
In the sorting network, the fixed KG order length $N$ is set to 8 and 10 for WebNLG and DART, respectively. The POS generator operates on the POS sequences parsed from ground-truth sentences via NLTK, and trains from BART-Base pretrained model~\cite{bart}. For Word-Encoder-Decoder, we follow the code in JointGT~\cite{jointgt} and utilize Bart-Base with self-attention~\cite{shaw2018self}. The beam search size for generating sentences in inference time is set to 5. We optimize all the parameters under the supervision of the total loss in Eq.~\ref{totalloss} using the OpenAI AdamW optimizer. The loss weights $\lambda_1$, $\lambda_2$ and $\lambda_3$ in total training loss (Eq.\ref{totalloss}) are set to 0.7, 0.4, and 0.3, respectively. 
\subsection{Main Results on WebNLG and DART}
	\begin{table}
		\centering
		\scalebox{0.6}{
			\begin{tabular}{lcccc}
				\toprule[0.8pt]
				\textbf{Datasets} & \multicolumn{4}{c}{WEBNLG} \\
				\midrule[0.5pt]
				\textbf{Metrics} & B-4 & R-L & CIDEr & Chrf++ \\
				\midrule[0.5pt]
				Li et al.$^\dagger$~\cite{fewshot} & {57.10} & 75.20 & {4.20} & {75.00} \\
				\cdashline{1-5}[1pt/1pt]
				Li et al.~\cite{fewshot} & \underline{61.88} & 75.74 & \textbf{6.03} & \underline{79.10} \\
			GraphWriter~\cite{graphtrans} & 45.84 & 60.62 & 3.14 & 55.53 \\
				CGE-LW~\cite{globallocal} & 48.60 & 62.52 & 3.85 & 58.66 \\
				T5-Base~\cite{plms} & 48.86 & 65.57 & 3.99 & 66.08 \\
				BART-Base~\cite{plms} & 49.81 & 63.10 & 3.45 & 67.65 \\
				T5-Large~\cite{plms} & 58.78 & 68.22 & 4.10 & 74.40 \\
				BART-Large~\cite{plms} & 52.49 & 65.61 & 3.50 & 72.00 \\
				JointGT$^\dagger$~\cite{jointgt} &57.00&	\underline{77.10}&	4.73&	76.90\\		
				S-OSC(ours) & \textbf{61.90} & \textbf{79.30} & \underline{5.30} & \textbf{79.70} \\
				\cdashline{1-5}[1pt/1pt]
                S-OSC(ours)-GT & 63.10 & 80.00 &5.40 & 81.00\\
				\bottomrule[0.8pt]
			\end{tabular}
		}
		\caption{Results for different models on WebNLG dataset. B-4 and R-L are short for BLEU-4 and ROUGE-L, respectively. \textbf{Bold} and \underline{underline} fonts represent the best and the second best performing results. 
		"S-OSC-GT" indicate our "S-OSC" model with ground-truth order in inference (The same term is applied in the following). $^\dagger$ denotes the results of reproduction. Other cited results are from Li et al.~\cite{fewshot}. }
		\label{tab:webnlg}
	\end{table}
We compare our S-OSC model with other state-of-the-art methods on WebNLG and DART. Results on WebNLG are shown in Table~\ref{tab:webnlg}. It can be observed from Table~\ref{tab:webnlg} that our S-OSC model outperforms all the previous methods in three evaluation metrics, i.e., B-4, R-L and Chrf++, except for the CIDEr value being in the second place compared to the best performing CIDEr result reported by the model in Li et al.~\cite{fewshot}. Note that our S-OSC model contains one learning-based sorting network supervised by the ground-truth order extracted from KG and the ground-truth sentence. 
We also report our model's results under the ground-truth order during inference time (denoted as "S-OSC-GT") which serves as the upper bound for our sorting performance. From the comparison of our model with predicted order "S-OSC" and our model with ground-truth order "S-OSC-GT", we can see that our model results with predicted order are close to the results with ground-truth order with the result gaps less than 1.2 points in most metrics, which shows the advantage of our learning-based sorting network for generating KG description order.
	\begin{table*}
		\centering		
		\scalebox{0.8}{
			\begin{tabular}{lccccc}
				\toprule[0.8pt]
				\textbf{Datasets} & \multicolumn{5}{c}{DART} \\
				\midrule[0.5pt]
				\textbf{Metrics} & B-4 & METEOR & MoverScore & BERTScore &BLEURT \\
				\midrule[0.5pt]
				Seq2Seq-Att~\cite{dartdataset} &29.66&	0.27&	0.31&	0.90&	-0.13\\
				End-to-End Transformer~\cite{ferreira2019neural} &27.24	&0.25 &	0.25	&0.89	&-0.29 \\
				T5-Large~\cite{plms} & {50.66} &	{0.40}	& \underline{0.54} & \underline{0.95} &	{0.44}\\
				BART-Large~\cite{plms} & 48.56&	0.39&	0.52&	\underline{0.95}&	0.41\\
                JointGT~\cite{jointgt} & \underline{54.24}&	\textbf{0.44}&\textbf{0.64}&	\textbf{0.96}&	\textbf{0.59}\\
                 S-OSC(ours)&\textbf{62.01}&	\underline{0.43}&\textbf{0.64}&\textbf{0.96}&	\underline{0.49}\\
                 \cdashline{1-5}[1pt/1pt]
                 S-OSC(ours)-GT& 64.53&	0.44&	0.66&0.96&	0.51\\
				\bottomrule[0.8pt]
			\end{tabular}
		}
		\caption{The results of different models on DART dataset.}
		\label{tab:dartexp}
	\end{table*}
	
	Results on DART are shown in Table~\ref{tab:dartexp}. 
	From previous models' results, we can see that pre-training on the large corpus (e.g., "T5-Large" and "Bart-Large") brings significant result improvement compared to "Seq2Seq-Att" and "End-to-End Transformer". Our S-OSC model further achieves new state-of-the-art results in all five metrics compared to these pre-training based models (e.g., "T5-Large") with B-4 score improved by 11.35 points, METEOR score improved by 0.03 points, Moverscore improved by 0.1 points, BERTScore improved by 0.01 points, and BLEURT improved by 0.05 points. JointGT outperforms all the previous baseline models. Compared with JointGT~\cite{jointgt}, our model can also obtain an improvement of 7.77 points in B-4 score. This result also validates the effectiveness of our model in improving the fluentness and authenticity of the generated sentences with the proposed learning-based sorting network and consistency enhancement under the POS syntactic and semantic context constraints. Similar to the results on WebNLG, our model S-OSC with predicted order can achieve the performance close to that with ground-truth order.
\subsection{Ablation Study}
In this section, we conduct extensive experiments on WebNLG to evaluate various factors in the word copy or prediction during sentence generation, e.g., the POS generator and semantic context (SC) scoring. We compare our full S-OSC model with the following variations: without word copy component and just use the predicted word from Word Decoder at each time step (w/o CP), without POS information in copy probability prediction (w/o POS), without semantic context scoring in copy probability prediction (w/o SC), without POS tag information and semantic context scoring in copy probability prediction (w/o POS and SC) and relying on the last hidden states to predict the copy probability as in previous methods~\cite{graphtrans,copypaste}. 
	\begin{table}
		\centering
		\scalebox{0.7}{
			\begin{tabular}{ccccc}
				\toprule[0.8pt]
				\textbf{Methods} & B-4 & R-L & CIDEr & Chrf++ \\
				\midrule[0.5pt]
				S-OSC &63.10&80.0&5.40&81.0 \\
				w/o CP & 59.15 & 77.8 & 4.99 & 78.0 \\
				w/o POS and SC  & 59.80 & 77.7 & 5.01 & 78.1 \\
				w/o POS & 61.00 & 78.5 & 5.10 & 78.6 \\
				w/o SC & 61.06 & 78.9 & 5.21 & 79.1 \\
				\bottomrule[0.8pt]
			\end{tabular}
		}
		\caption{Ablation analysis for copy or prediction component in our model on WebNLG. S-OSC here is under the ground-truth order.}
		\label{tab:removemodul}
	\end{table}
	From the results in Table~\ref{tab:removemodul}, we can observe that:
	(1) By removing the word copy component and directly taking the predicted word from Word Decoder at each time step (w/o CP), the results drop significantly by 3.95 points in B-4, 2.2 points in R-L, 0.41 points in CIDEr and 3 points in Chrf++.
	(2) We then consider incorporating the copy prediction component, but just rely on the last hidden states to predict the copy probability as in previous methods~\cite{graphtrans,copypaste} without POS tag information and semantic context scoring in copy probability prediction (w/o POS and SC). The results improve slightly in all the metrics compared with that of "w/o CP".
	(3) We then evaluate the effectiveness of POS syntactic information and semantic context scores in improving the quality of the generated sentences by removing each of them at a time.
	
	Compared to full S-OSC model, without POS information in copy probability prediction (w/o POS), the results drop by 2.1 points in B-4, 1.5 points in R-L, 0.3 points in CIDEr and 2.4 points in Chrf++.
	Compared to full S-OSC model, without semantic context scoring in copy probability prediction (w/o SC), the results drop by 2 points in B-4, 1.1 points in R-L, 0.2 points in CIDEr and 1.9 points in Chrf++. Both "w/o POS" and "w/o SC" improve consistently compared to previous copy policy in "w/o POS and SC".

	To further reveal more details about our model, we conduct extra ablation studies regarding the following questions.
	(1) How does triplet structure encoding in the sorting network help the model compared with direct node encoding without triplet structure context?
	(2) How does our model perform for different triplet numbers in KG?
	(3) How does the sliding window size in semantic context scoring function affect the model performance?

 \paragraph{(1) Triplet structure encoding in sorting network.}
 To show the effect of triplet structure encoding (triple-level) in our sorting network, we investigate the performance of direct node encoding without triplet structure (node-level), as well as our S-OSC model's results with random sorting order. Results on WebNLG are shown in Table~\ref{tab:differentwebnlg} and the upper bound result is shown with ground-truth order (GT).
 \begin{table}
	\centering
	\scalebox{0.7}{
		\begin{tabular}{ccccc}
			\toprule[0.8pt]
			\textbf{Methods} & B-4 & R-L & CIDEr & Chrf++ \\
			\midrule[0.5pt]
			{GT} & 63.10 & 80.0 & 5.4 & 81.0\\
			\cdashline{1-5}[1pt/1pt]
			\textbf{TS} & 61.90 & 79.3 & 5.3 & 79.7 \\
			NS & 60.00 & 77.0 & 5.1 & 78.0 \\
			RS  & 55.20 & 72.0 & 4.7 & 76.0 \\
			\bottomrule[0.8pt]
		\end{tabular}
	}
	\caption{The results of sorting order on WebNLG. TS denotes the triple-level sorting order, NS is the node-level sorting order~\cite{fewshot} and RS represents the random sorting order. The same symbols are as below.}
	\label{tab:differentwebnlg}
\end{table}
 From Table~\ref{tab:differentwebnlg}, sorting with triple-level sorting order "TS" outperforms random sorting order "RS" significantly, and also outperforms node-level sorting order "NS" by 1.9 points in B-4, 2.3 points in R-L, 0.2 points in CIDEr and 1.7 points in Chrf++, showing the advantage of triplet structure encoding compared to node encoding without triplet structure context. The results on DART dataset show similar trend and are reported in supplementary material.
\paragraph{(2) Different Knowledge Graph Sizes.}
\begin{figure}
		\centering
		\includegraphics[scale=0.24]{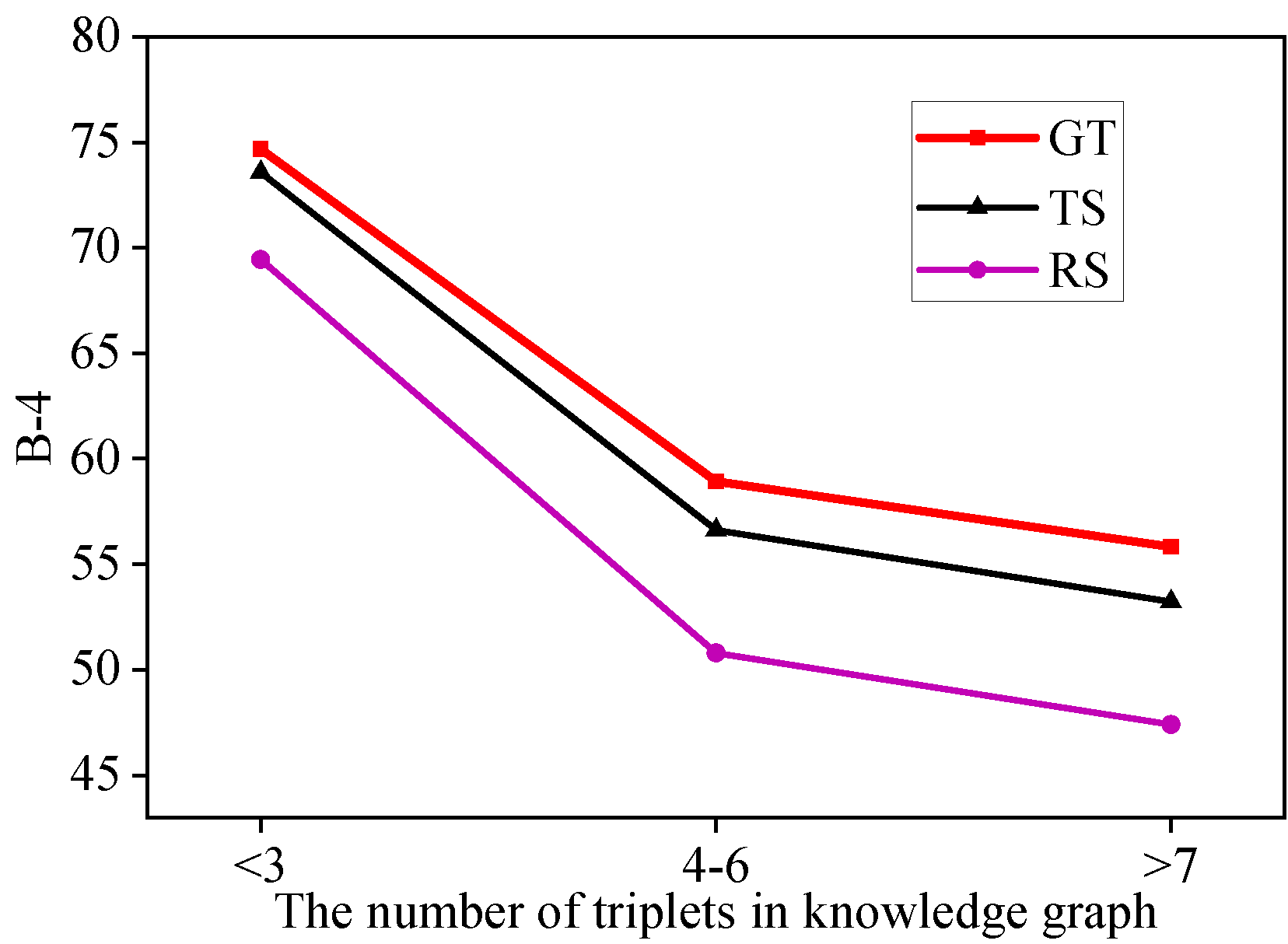}
		\caption{The B-4 results in three groups with different KG sizes on DART dataset for different sorting orders.}
		\label{fig:differentKG}
\end{figure}
To verify the effect of our S-OSC model performance on different KG sizes, we split the test dataset into three subsets according to the size of the KG, i.e., the number of triples in KG is less than 3, between 4 and 6, and more than 7, and reports results on each subset. DART is selected for experiment due to its wide distribution of KG sizes. The results of B-4 in each subset are plotted in Figure~\ref{fig:differentKG}, showing that as the number of triples in KG increases, the difficulty of sorting and KG-to-text generation increases and the B-4 results generally decrease in all three methods ("GT", "RS" and "TS"). Also, our model's superiority exhibits significant improvement in all three subsets with different KG sizes (comparing "TS" with "RS"). Other metric results show similar trend and are reported in supplementary material.
\paragraph{(3) Sliding window size in the semantic context scoring.}
We plot the B-4 results for different sliding window sizes of semantic context scoring function on WebNLG dataset, shown in Figure~\ref{fig:slddingwindows}. From Figure~\ref{fig:slddingwindows}, the model achieves the best performance when the sliding window size is 3. Other metric results show similar trend and are reported in supplementary material. Thus, we set the sliding window size to 3 in the experiments.
 \begin{figure}
		\centering
		\includegraphics[scale=0.24]{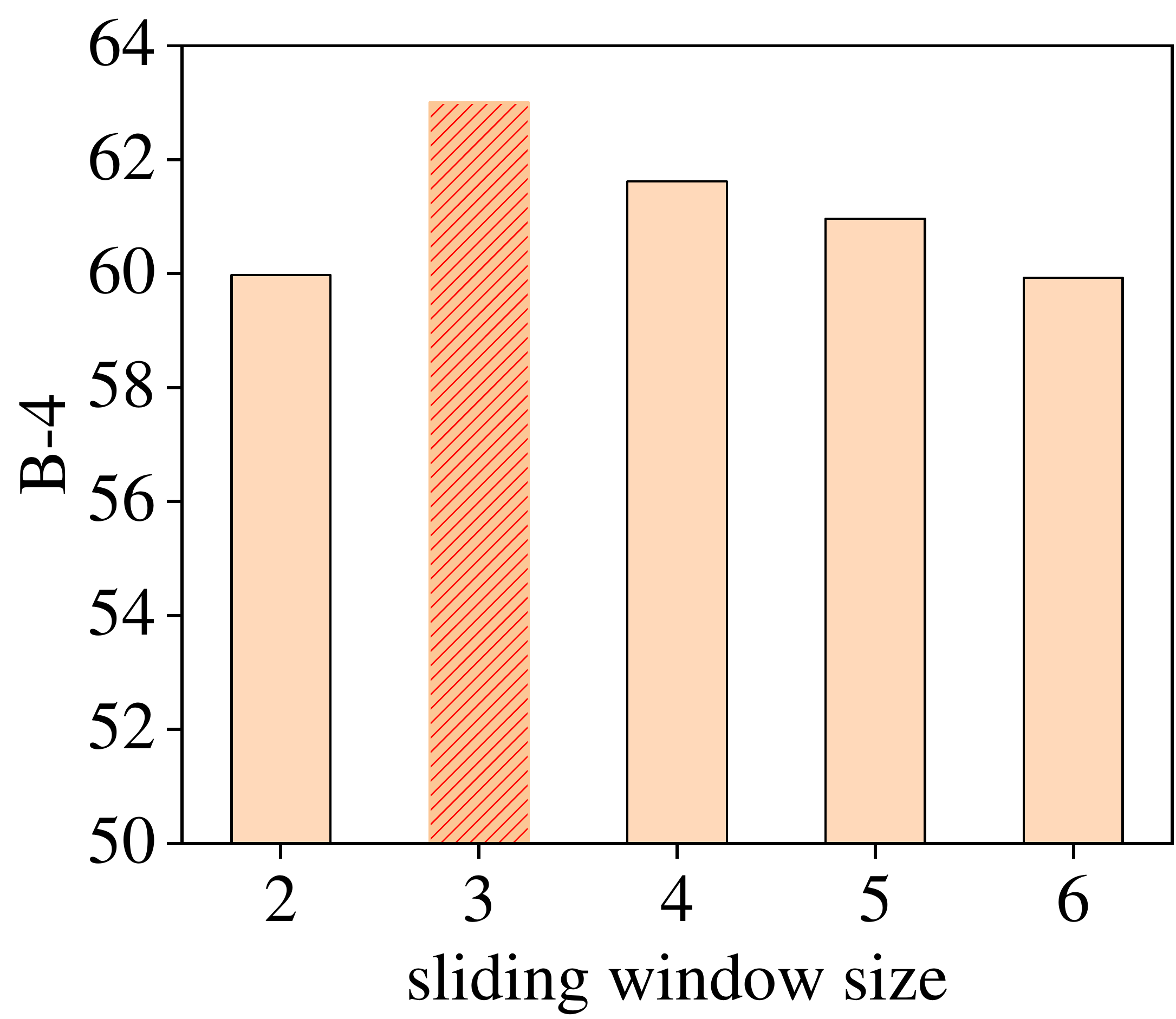}
		\caption{The B-4 results of different sliding windows on WebNLG dataset.}
		\label{fig:slddingwindows}
	\end{figure}
\subsection{Human Evaluation}
	\begin{figure*}[htbp]
		\centering
		\includegraphics[scale=0.56]{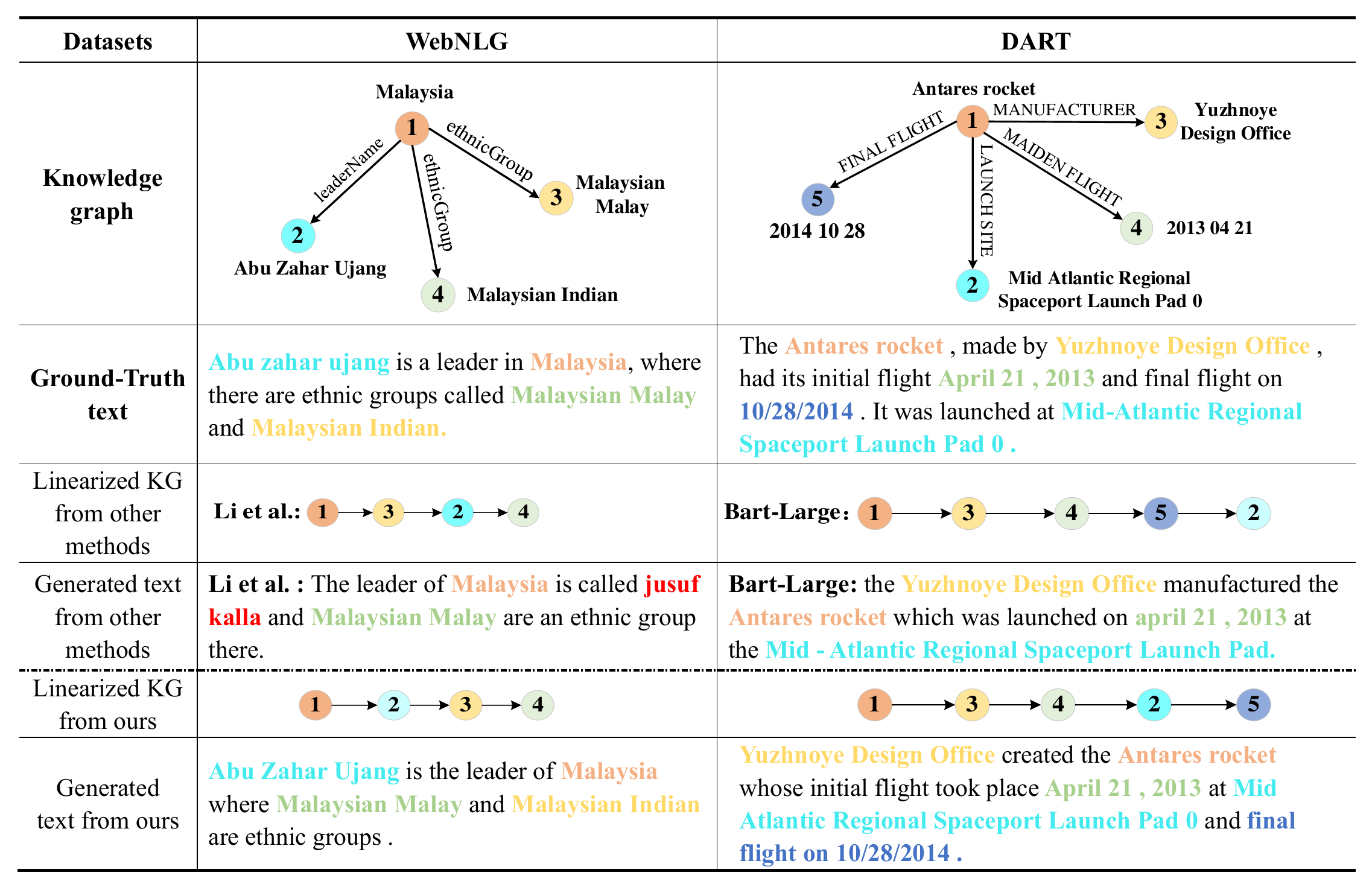}
		\caption{Two examples from WebNLG and DART. The text in color is corresponding to the KG nodes. The generated texts in red color indicate that the generated texts contract with KG.}
		\label{fig:quality}
	\end{figure*}
We conduct the human evaluation on WebNLG to further evaluate the generated text. In this paper, we adopt the same human evaluation criteria as Chen et al.~\cite{chen2019few}, i.e., Factual correctness including Supp (counting the number of facts that co-exist in the KG and generated text) and Cont (counting the facts in the generated texts missing from or contradicting with KG), Language naturalness including NF (evaluating the accuracy and fluentness of generated sentences). In addition to using absolute score at 5-point metric in NF, we also use relative ranking scores (termed NA).	We randomly select 100 knowledge graphs for human evaluation. Five native English speakers volunteer to score all the 100 knowledge graphs. Table~\ref{tab:humanevlu} reports the results\footnote{ Cohen's kappa coefficients for the first two factors are 0.79, 0.83,0.75,0.76.}. We can observe that: the generated texts of our S-OSC model are more authentic and consistent with KG than the method in Li et al.~\cite{fewshot}.
	\begin{table}
	\centering
	\scalebox{0.7}{
		\begin{tabular}{ccccc}
			\toprule[0.8pt]
			\textbf{Methods} & Supp.\textuparrow & Cont.\textdownarrow &NF\textuparrow & NA\textuparrow   \\
			\midrule[0.5pt]
			\textit{ground-truth} &{3.82} &{0.10} &{4.75} &{2.93}\\
			Li et al.\cite{fewshot} &3.48 &0.33 &3.7 &2.10\\
			S-OSC &\underline{3.77} &\underline{0.15} &\underline{4.25} &\underline{2.48}\\
			\bottomrule[0.8pt]
		\end{tabular}
	}
	\caption{The results of human evaluation on WebNLG dataset.}
	\label{tab:humanevlu}
\end{table}

\subsection{Qualitative Analysis}
We show two qualitative examples from WebNLG and DART in Figure~\ref{fig:quality}. From the example of WebNLG, we can see that the previous method in Li et al.~\cite{fewshot} generates the wrong description order with node 3 and node 2 exchanged positions. This causes the generated sentence from Li et al.~\cite{fewshot} containing the wrong text "jusuf kalla" noted in red color and missing the text "Malaysian Indian", while our S-OSC method generates the right order and consistent sentences. In the example from DART, though our S-OSC model generates relative inferior order with node 2 and node 5 exchanged positions, relying on our strong sentence generator with syntactic constraint and semantic consistency constraint, our model is able to generate semantic consistent sentence to the KG with the right words copied from KG. However, in this example, Bart-Large~\cite{plms} still misses some key word description from KG, e.g., "the final fight on 10/28/2014" and "Launch Pad 0", though conditioned on the right predicted order, showing the inferior performance of their copy or prediction module in sentence generation.

\section{Conclusion}
This paper proposes a learning-based sorting network to obtain the optimal description order for KG-to-text generation. Additionally, our model incorporates POS generator and semantic context scoring to selectively copy words from KG and improve the word authenticity in generated sentences. Extensive experiments show that our model outperforms previous state-of-the-art approaches. In the future, we will introduce casual inference into the model to further improve the reasoning ability.

\bibliography{anthology,custom}
\bibliographystyle{acl_natbib}

\end{document}


\maketitle
\section{0-1 Generator}\label{01gen}
\begin{figure}
		\centering
		\includegraphics[scale=0.7]{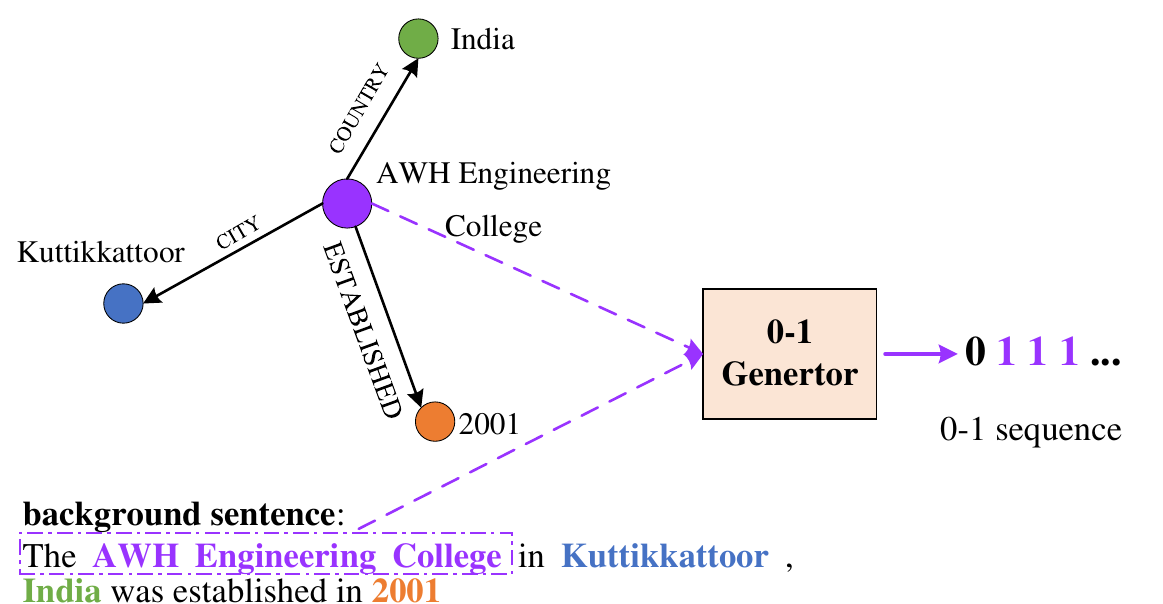}
		\caption{The steps of obtaining the 0-1 sequence. When the KG node exists in the text, all mentions of the node are masked as 1, otherwise 0. }
		\label{fig:append01}
	\end{figure}

	In order to make the generated text more consistent with the KG, we use a 0-1 generator to obtain the ground-truth of copy or prediction order, where 0 represents a token from generation and 1 from KG. The steps are as shown in Figure~\ref{fig:append01}.
\section{Ground-Truth Sorting Order}\label{groundtruth}
	The description sorting order is essential for the generation process. In this paper, we sort the order according to the triple-level. Taking Figure~\ref{fig:append01} as an example, we can obtain three triplets, i.e., (AWH Engineering College, COUNTRY, India), (AWH Engineering College, ESTABLISED, 2001) and (AWH Engineering College, CITY, Kuttikkattoor). In the reference sentence, the triple (AWH Engineering College, CITY, Kuttikkattoor) appears first, (AWH Engineering College, COUNTRY, India) appears second and (AWH Engineering College, ESTABLISED, 2001) last. Thus, we can form the order 2,0,1 for the providing KG. Note that, we only record the position of the triplet where it first appeared regardless of many times.    
\section{Code Re-implementation}
We re-implement the author's code in Li et al.~\cite{fewshot} as our baseline. It can be observed from Table~\ref{tab:webnlg} that our S-OSC model outperforms the baseline and the results of Li et al. reported from their paper in three evaluation metrics,B-4, R-L and Chrf++ metrics. Although Li et al. achieves the best CIDEr score in their reported results, this score is not replicable and our S-OSC still obtains 1.1 point higher than that of the replicated baseline.  
\begin{table}
		\centering
		\scalebox{0.7}{
			\begin{tabular}{lcccc}
				\toprule[0.8pt]
				\textbf{Datasets} & \multicolumn{4}{c}{WEBNLG} \\
				\midrule[0.5pt]
				\textbf{Metrics} & B-4 & R-L & CIDEr & Chrf++ \\
				\midrule[0.5pt]
				Baselines$^{\dagger}$ & 57.00 & 75.20 & 4.20 & 75.00 \\	
				Li et al.\cite{fewshot} & \underline{61.88} & \underline{75.74} & \textbf{6.03} & \underline{79.10} \\
				S-OSC(ours) & \textbf{61.90} & \textbf{79.30} & \underline{5.30} & \textbf{79.70} \\
				\bottomrule[0.8pt]
			\end{tabular}
		}
		\caption{Results for different models on WebNLG dataset. B-4 and R-L are short for BLEU-4 and ROUGE-L, respectively. Baseline is the reproduced results of Li et al.~\cite{fewshot}. \textbf{Bold} and \underline{underline} represent the best and the second best performing results. (the same term is used below)}
		\label{tab:webnlg}
	\end{table}	

\section{Triplet Structure Encoding in Sorting Network on DART~\cite{dartdataset}}
 From Table~\ref{tab:differentdart}, encoding with triple-level sorting order "TS" outperforms random sorting order "RS" significantly, and also outperforms node-level sorting order "NS" by 1.35 points in B-4, 0.01 points in METEOR, 0.02 points in BERTScore and 0.13 points in BLEURT, showing the advantage of triplet structure encoding compared to node encoding without triplet structure context.
	\begin{table*}
		\centering
		\scalebox{0.9}{
			\begin{tabular}{c|ccccc}
				\toprule[0.8pt]
				\textbf{Methods} & B-4 & METEOR & MoverScore & BERTScore &BLEURT \\
				\midrule[0.5pt]
				GT &64.53 &	0.44&	0.66&	0.96&	0.51\\
				TS & 62.01	&0.43&	0.64&	0.96&	0.51 \\
				NS & 60.66&	0.42&	0.64&	0.94&	0.38	 \\
				RS  & 58.67&	0.42&	0.62&	0.93&	0.34	\\
				\bottomrule[0.8pt]
			\end{tabular}
		}
		\caption{The results of sorting order on Dart dataset.}
		\label{tab:differentdart}
	\end{table*}

\begin{table}[t!]
	\centering
	\scalebox{0.8}{
		\begin{tabular}{c|cccc}
			\toprule[0.8pt]
			\textbf{Methods} & B-4 & R-L & CIDEr & Chrf++ \\
			\midrule[0.5pt]
			With local POS & 63.10 & 80.00 & 5.40 & 81.0 \\
			With global POS & 61.28 & 79.32 & 5.28 & 79.2 \\
			\bottomrule[0.8pt]
		\end{tabular}
	}
	\caption{Performance comparisons of our model with different POS information (local POS and global POS) on WebNLG dataset.}
	\label{tab:poslocalglobal}
\end{table}
	
\begin{sidewaystable}[thp]
		\centering
		\resizebox{\textwidth}{!}
		{
			\begin{tabular}{c|cccc|cccc|cccc}
				\toprule[0.8pt]
				\textbf{Datasets} & \multicolumn{4}{c|}{<3} & \multicolumn{4}{c|}{4-6} &\multicolumn{4}{c}{>7}\\
				\midrule[0.5pt]
				\textbf{Metrics} & METEOR & MoverScore & BERTScore &BLEURT  & METEOR & MoverScore & BERTScore &BLEURT & METEOR & MoverScore & BERTScore &BLEURT\\
				\midrule[0.5pt]
				{GT} &0.51&	0.75&	0.97&	0.61&
				0.41&	0.57&	0.94&	0.40&
				0.40&	0.53&	0.94&	0.39\\
				TS 	&0.50	&0.74	&0.96	&0.60
				&0.40	&0.56	&0.94	&0.40
				&0.39	&0.52	&0.94	&0.38\\	
				NS 	&0.49	&0.72	&0.97	&0.58
				&0.39	&0.54	&0.94	&0.37
				&0.38	&0.49	&0.94	&0.38	 \\	
                RS 	&0.47	&0.71	&0.97	&0.57	&0.38	&0.52	&0.93	&0.36	&0.37	&0.47&0.93&0.35 \\
				\bottomrule[0.8pt]
			\end{tabular}
		}
		\caption{The results (METEOR, MoverScore, BERTScore and BLEURT metrics) of different sorting orders in the three subsets with difference KG sizes on Dart dataset.}
		\label{tab:dartdifferenttable}
\end{sidewaystable}
\section{Different Knowledge Graph Sizes}
Table~\ref{tab:dartdifferenttable} shows that when the number of triples in KG increases, the difficulty of sorting and KG-to-text generation increases (see "GT" and "TS") and our model's superiority exhibits more relative improvement (comparing "TS" with "NS"). When the size of knowledge graphs is lower than three, these orders have similar performances. This is mainly due to the fact that less triplets have less effect for the model. However, "TS" significantly outperforms "NS" and "RS". Moreover, as the results reported in Table~\ref{tab:dartdifferenttable}, "TS" consistently outperforms "RS" and "NS" in terms of METEOR and MoverScore metrics, and obtains similar performance to "GT" in BERTScore and BLEURT metrics.
\section{Sliding Window Size in Semantic Context Scoring}
\begin{figure}
	\centering
	\includegraphics[scale=0.3]{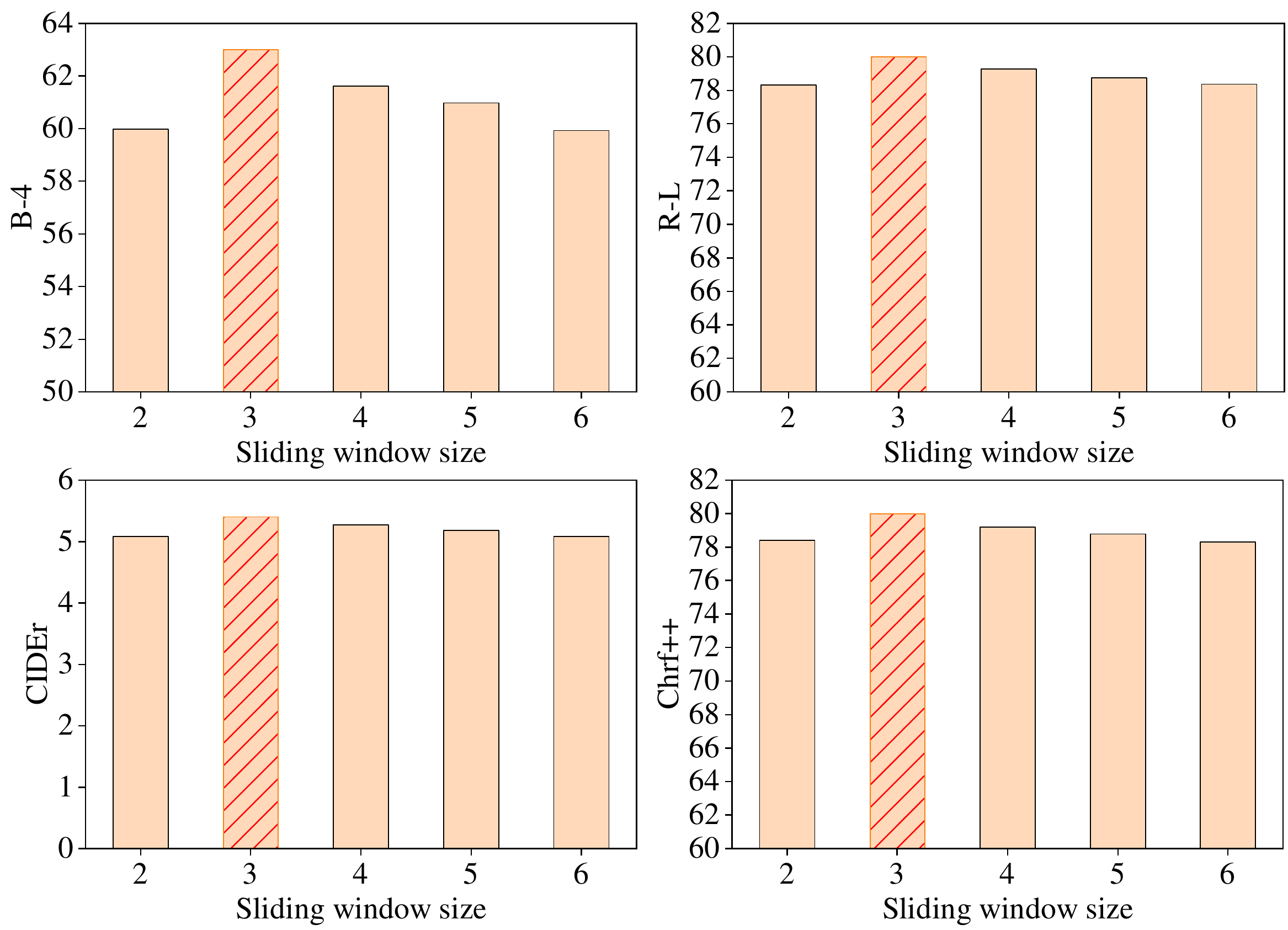}
	\caption{The results of different sliding windows on WebNLG dataset under B-4, R-L, CIDEr and CHrf++ metrics.}
	\label{fig:slddingwindows}
\end{figure}

As the B-4 result reported in the main paper, the other metric results in Figure~\ref{fig:slddingwindows} also show that the model achieves the best performance when the sliding window size is 3.  
\section{Global and Local POS Information}
We further introduce two forms of POS tag information, i.e., the global tag information (the final last hidden state of POS generator) and local tag information (the last hidden states of POS generator at each time step). The results of the comparison are reported in Table~\ref{tab:poslocalglobal}. 
	
We can observe that the model equipping with local POS information performs better than that of equipping with global information. This is mainly because the local POS information can provide more fine-grained syntactic information. Thus, it can further ensure the authenticity of the generated sentences.

\bibliography{anthology,custom}
\bibliographystyle{acl_natbib}